\documentclass[letterpaper, 10 pt, conference]{ieeeconf}
\IEEEoverridecommandlockouts    
\usepackage{graphics}           
\usepackage{times}              
\usepackage{amsmath}            
\usepackage{amssymb}            
\usepackage{graphicx}
\usepackage{algorithm}
\usepackage[noend]{algpseudocode}
\usepackage{booktabs}
\usepackage{color}
\definecolor{instructioncolor}{rgb}{.5,.5,.5}

\usepackage[font=small]{caption}

\def\eqref#1{Eq.~(\ref{#1})}

\makeatletter
\usepackage{xspace}
\DeclareRobustCommand\onedot{\futurelet\@let@token\@onedot}
\def\@onedot{\ifx\@let@token.\else.\null\fi\xspace}

\def\etal{{et al}\onedot}
\makeatother

\usepackage{array}
\newcolumntype{L}[1]{>{\raggedright\let\newline\\\arraybackslash\hspace{0pt}}m{#1}}
\newcolumntype{C}[1]{>{\centering\let\newline\\\arraybackslash\hspace{0pt}}m{#1}}
\newcolumntype{R}[1]{>{\raggedleft\let\newline\\\arraybackslash\hspace{0pt}}m{#1}}

\usepackage[table]{xcolor}
\usepackage{multirow}
\usepackage{hyperref}
\hypersetup{
    colorlinks=true,
    citecolor=blue,
    linkcolor=magenta,
    filecolor=magenta,      
    urlcolor=magenta,
}

\title{\LARGE \bf Image-Goal Navigation Using Refined Feature Guidance \\ and Scene Graph Enhancement}

\author{Zhicheng Feng \and Xieyuanli Chen \and Chenghao Shi \and Lun Luo \and Zhichao Chen \and Yun-Hui Liu \and Huimin Lu
	\thanks{Z. Feng, X. Chen, C. Shi and H. Lu are with the College of Intelligence Science and Technology, and the National Key Laboratory of Equipment State Sensing and Smart Support, National University of Defense Technology, China. L. Luo is with Zhejiang University, China. Z. Chen is with Jiangxi University of Science and Technology, China. Y.H. Liu is with the T Stone Robotics Institute and Department of Mechanical and Automation Engineering, the Chinese University of Hong Kong, China.}
    \thanks{Corresponding author: Huimin Lu (lhmnew@nudt.edu.cn).}%
	\thanks{This work was supported in part by the National Science Foundation of China (Grant No. 62403478, U22A2059 and 62203460), Young Elite Scientists Sponsorship Program by CAST (No. 2023QNRC001), and  Major Project of Natural Science Foundation of Hunan Province (Grant No. 2021JC0004).
	}%
}

\begin{document}
	\maketitle
	\thispagestyle{empty}
	\pagestyle{empty}

	\begin{abstract}
        In this paper, we introduce a novel image-goal navigation approach, named RFSG. Our focus lies in leveraging the fine-grained connections between goals, observations, and the environment within limited image data, all the while keeping the navigation architecture simple and lightweight. To this end, we propose the spatial-channel attention mechanism, enabling the network to learn the importance of multi-dimensional features to fuse the goal and observation features. In addition, a self-distillation mechanism is incorporated to further enhance the feature representation capabilities. Given that the navigation task needs surrounding environmental information for more efficient navigation, we propose an image scene graph to establish feature associations at both the image and object levels, effectively encoding the surrounding scene information. Cross-scene performance validation was conducted on the Gibson and HM3D datasets, and the proposed method achieved state-of-the-art results among mainstream methods, with a speed of up to 53.5 frames per second on an RTX3080. This contributes to the realization of end-to-end image-goal navigation in real-world scenarios. The implementation and model of our method have been released at: \href{https://github.com/nubot-nudt/RFSG}{https://github.com/nubot-nudt/RFSG}.
	\end{abstract}

\section{Introduction}
\label{sec:intro}
Traditional navigation methods~\cite{thrun2002probabilistic,mur2015orb} typically execute navigation tasks through a sequential process, mainly relying on Simultaneous Localization and Mapping (SLAM) techniques. These methods first perceive environmental information to construct a map, followed by path planning from the current location to the target position using the generated map. In contrast, recent advancements in Image-Goal Navigation~\cite{lin2023learning,majumdar2022zson,sun2023fgprompt} have shifted focus from this modular approach towards developing end-to-end navigation solutions.
Image-goal navigation represents a paradigm shift in autonomous navigation, aiming to search targets without precise positional information, relying solely on a goal image~\cite{zhu2017target,krantz2023navigating}. These approaches present unique challenges: 1) Unlike conventional navigation systems that utilize accurate positioning data, image-goal navigation requires establishing robust feature associations between the goal image and real-time observations. 2) Current methods mostly focus on directly associating scene features or pre-constructing scene image associations in a database-like form, hardly combining the fine-grained utilization of the scene images and the simplicity of navigation architecture.
	
\begin{figure}[t]
	\centering
	\includegraphics[width=0.9\linewidth]{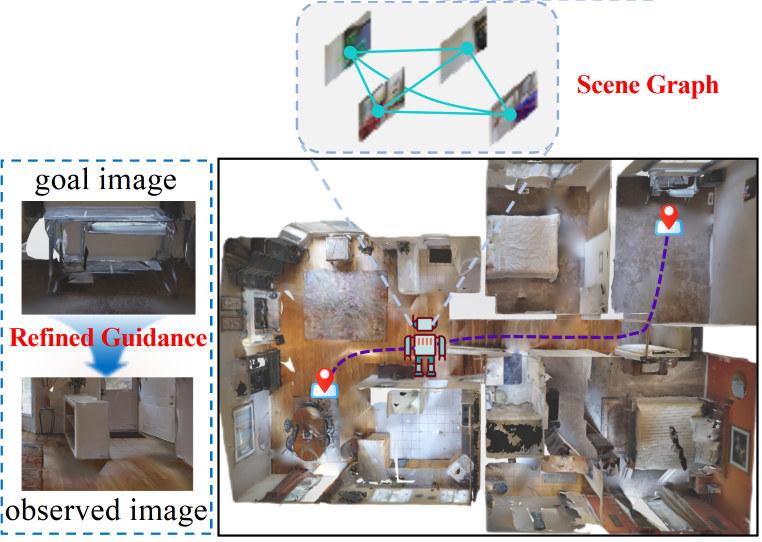}
	\caption{A robot relies only on the goal and observed images to accomplish navigation in the environment. We focus on building a fine-grained feature guidance from the goal to the observed images, and capturing important environmental information from multiple scene images through a scene graph. }
	\label{fig:motivation}
\end{figure}

To address the aforementioned challenges, we propose a novel approach that achieves state-of-the-art navigation performance and cross-scene generalization ability, while maintaining a lightweight architecture, as shown in Fig.~\ref{fig:motivation}. Our method employs a lightweight dual-branch encoder to process both the goal and observation images. To establish more robust associations between goal and observation features, we introduce a fine-grained feature fusion strategy that enables the effective integration of intermediate features in the backbone.

Additionally, inspired by knowledge distillation theory~\cite{zhang2019your}, we design a parameter-free self-distillation mechanism to encourage the shallow network model to emulate the weights of the deeper network, thereby further enhancing the representational capacity of the extracted features. To uncover the underlying relationships between environmental information and navigation actions, we construct an image-object scene graph that captures comprehensive environmental information.

In summary, the contributions of this paper are threefold:
\begin{itemize}
    \item We propose a novel image-goal navigation approach, called RFSG, using Refined Feature guidance and Scene Graph enhancement, achieving state-of-the-art navigation performance and cross-scene generalization ability, while maintaining a lightweight architecture.
    \item We introduce a novel feature fusion strategy to effectively combine goal and observation features. To further improve feature representation, we propose a parameter-free knowledge self-distillation technique to optimize the shallow network.
    \item We propose a scene graph that embeds both image and instance-level object features, effectively establishing dependencies among different images to provide rich environmental information for navigation.
\end{itemize}

\begin{figure*}[t]
    \centering
    \includegraphics[width=0.9\linewidth]{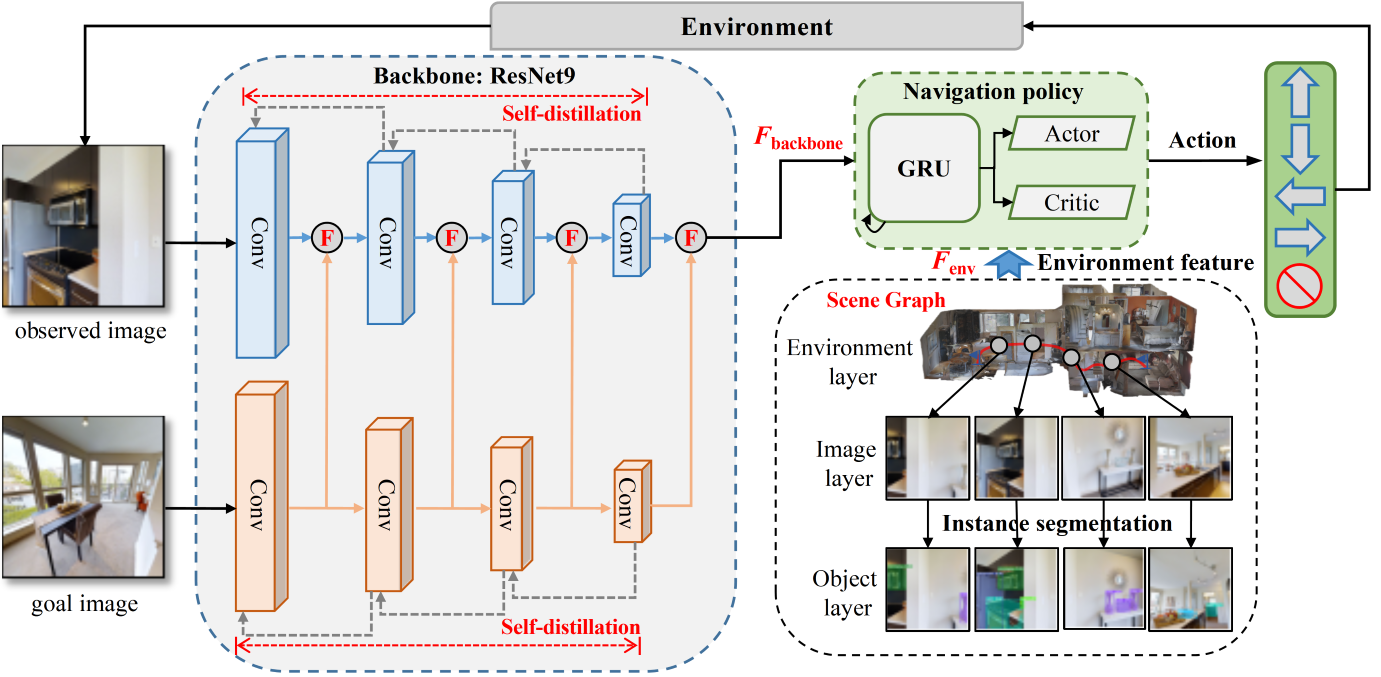}
    \caption{Overall structure of the proposed RFSG. Initially, feature extraction is achieved by the backbone ResNet9 with self-distillation, and a feature fusion strategy \textcircled{F} is introduced to fine-grained capture important goal and observation features. Subsequently, it combines the instance targets in the image to construct a scene graph to capture the environment information. Finally, the scene graph feature $F_\text{env}$ is embedded into the backbone feature $F_\text{backbone}$, and the next action of the robot is generated based on the Actor-Critic architecture. The actions of the robot contain FORWARD, BACKWARD, LEFT TURN, RIGHT TURN, and STOP.}
    \label{fig:method}
\end{figure*}

\section{Related Work}
\label{sec:related}

\textbf{SLAM-based Methods: }Most navigation methods modularize the entire architecture~\cite{thrun2002probabilistic}, which often relies on SLAM technology. By meticulously designing multiple modules in the SLAM system~\cite{mur2015orb}, navigation problems in specific environments can be solved. With the development of deep learning, some studies have replaced traditional modules with learning-based methods, such as using pose estimation models for localization~\cite{chaplot2018active} or constructing explicit or implicit maps~\cite{gupta2017cognitive,savinov2018semi,chaplot2020neural} based on neural networks. Meanwhile, some studies have completely replaced traditional SLAM with deep learning models, but this requires a large amount of annotated data and still depends on precise sensors~\cite{chaplot2020learning,avraham2019empnet}. Although SLAM-based methods can combine visual, inertial, LiDAR, and other data to ensure the reliability and accuracy of the navigation system, the complex coupling of multi-sensor systems also brings difficulties to their practical applications~\cite{parisotto2018neural, zhang2017neural}. Therefore, in the real world, navigation architectures based on direct image input may better meet the needs of the robot.

\textbf{RL-based Methods: }For the image-goal navigation task, a mainstream architecture is to predict the robot's next action based on Reinforcement Learning (RL)~\cite{al2022zero,sun2023fgprompt}, thereby planning the path from the current position to the goal position. Li~\etal~\cite{zhu2017target} are the first to solve the image-goal navigation problem in an end-to-end form, training the robot through a twin Actor-Critic architecture. Subsequently, some methods encouraged the robot to explore the environment by designing different reward strategies~\cite{lin2023learning,majumdar2022zson,du2021curious}. Al-Halah~\etal~\cite{al2022zero} design a novel reward mechanism to achieve a semantic search strategy. To tackle limited RGB image information, some methods combine different data with RGB~\cite{hahn2021no,ye2024rgbd} or provide additional information by constructing an extra memory storage unit~\cite{mezghan2022memory,kim2023topological,kwon2021visual}. Yadav~\etal~\cite{yadav2023offline,yadav2023ovrl} achieve navigation by jointly using GPS data and image features, effectively compensating for the lack of positional information in RGB. Ye~\etal~\cite{ye2024rgbd} construct a topological graph based on RGBD images to overcome the offset of the robot's relative pose. Mezghani~\etal~\cite{mezghan2022memory} construct a full visual memory unit, obtaining important memory features through an attention mechanism. Kim~\etal~\cite{kim2023topological} build a topological graph structure with input images as nodes, enhancing the robot's perceptual ability by remembering the observation objects of each node. These methods focus on providing more information to the robot by inputting additional data or constructing complex memory units, which makes the overall architecture often complex and pays less attention to how to exploit the inherent fine-grained features of the goal and observation images.

\section{Our Approach}
\label{sec:main}
   
The image-goal navigation task aims to guide a robot to a target location represented by a goal image through iterative action planning based on observed images. To address this challenge, we propose RFSG, a novel framework that first combines intermediate goal and observation features using a fine-grained fusion strategy. Additionally, we introduce a self-distillation mechanism to enhance model performance while preserving the lightweight architecture, as detailed in Section~\ref{sec1}.
Furthermore, to capture a richer environmental context, we construct a scene graph that integrates both image-level and instance-level features, enabling comprehensive environmental information encoding, as described in Section~\ref{sec2}. Finally, the fine-grained features and scene graph features are jointly utilized to generate navigation policy through reinforcement learning, as outlined in Section~\ref{sec3}. The overall architecture of our proposed approach is illustrated in Fig.~\ref{fig:method}.
    
\begin{figure}[tbh]
    \centering
    \includegraphics[width=1\linewidth]{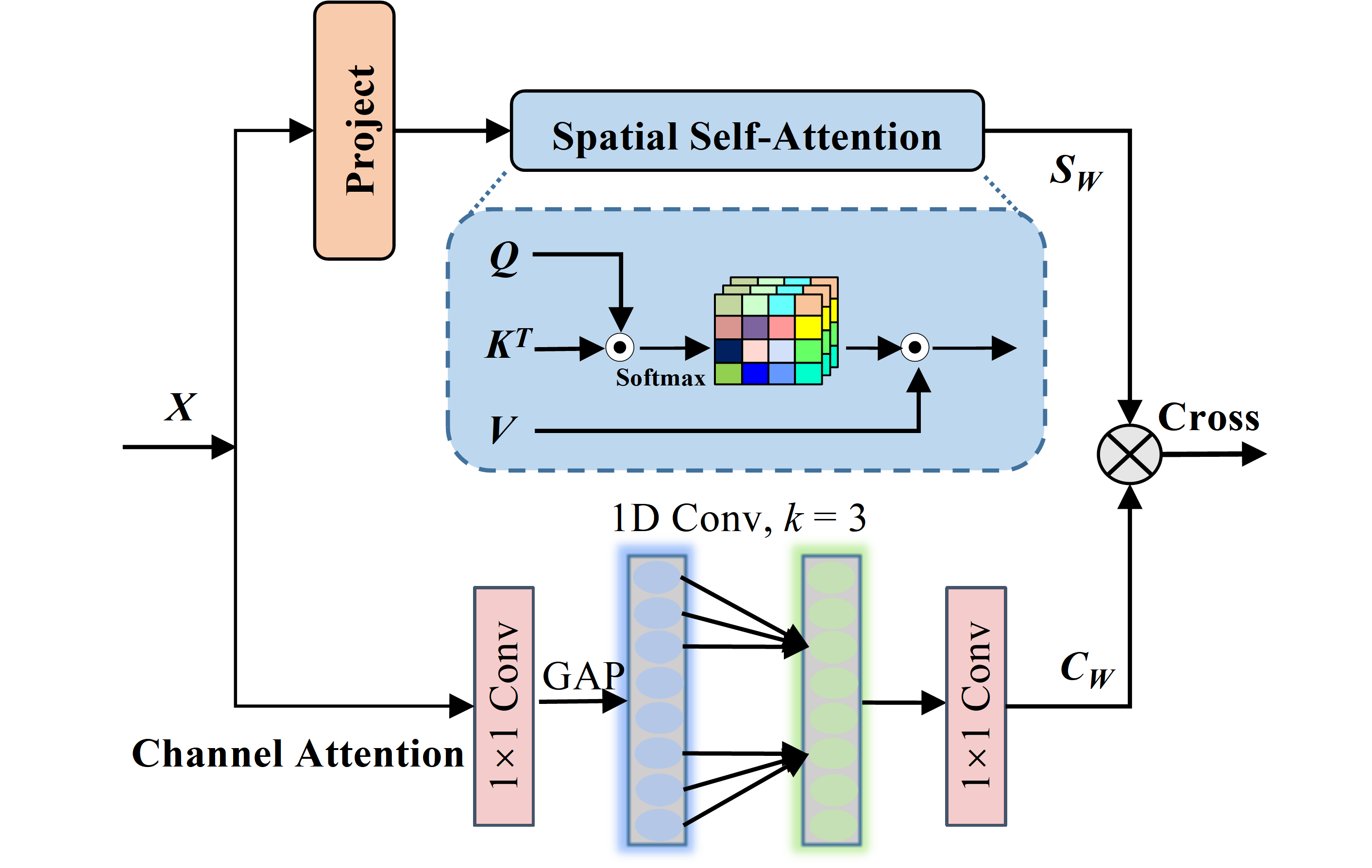}
    \caption{The proposed spatial-channel attention. Important features are focused on the spatial and channel dimensions respectively, and then cross-multiplication $\otimes$ is utilized to obtain the output feature. Here, $\odot$ is the matrix multiplication.}
    \label{fig:SCAM}
\end{figure}

\subsection{Feature Extraction and Fusion Module}\label{sec1}
In this module, we aim to effectively utilize both the goal and observed images, extracting joint features for navigation policy generation. Following the previous works~\cite{chaplot2020learning,al2022zero,sun2023fgprompt}, we employ ResNet9~\cite{he2016deep} as the backbone network for fast feature extraction from both the goal and observed images. Subsequently, we propose a refined feature fusion strategy to establish robust associations between the goal and observed images. To further enhance the representational capacity of the features, we propose to employ a parameter-free self-distillation mechanism that encourages the shallow network layers to approximate the weight distribution of deeper layers, thereby improving the backbone's feature extraction performance without introducing additional learnable parameters. The detailed implementation of the feature fusion strategy and self-distillation mechanism is presented in the following sections.
    
\subsubsection{Feature Fusion}
Following existing methods \cite{jang2022bc,yu2022using,sun2023fgprompt,brohan2022rt}, we adopt the same paradigm to fuse goal image feature $X_\text{goal}$ with observed image feature $X_{\text{observe}}$ using learnable weights $W_1$ and $W_2$, given as follows: 
\begin{equation}
    Y = W_1 \cdot X_{\text{observe}} \cup W_2 \cdot X_{\text{goal}},
    \label{con:1}
\end{equation}
where $\cup$ represents the summing operation. To capture distinctive features from the images, we employ a Spatial-Channel Attention mechanism along with a Weight Decoupling Module to generate weights.
	
\textbf{Spatial-Channel Attention: } 
To capture the refined features of the goal image, we establish attention in spatial and channel dimensions, respectively. Specifically, we propose Spatial-Channel Attention~(SCA) that combines the multi-head spatial attention mechanism with the local-global channel attention mechanism, as shown in Fig.~\ref{fig:SCAM}. 

In spatial attention, the input feature tensor undergoes an initial transformation through a convolutional operation. This transformed representation is subsequently partitioned into query ($Q$) and key ($K$) tensors along the channel axis through dimensional factorization. Given that the goal feature requires concentrated contextual modeling, the value ($V$) tensor maintains the original input feature. To enhance multi-scale spatial modeling capabilities, each channel dimension is conceptualized as an independent attention head, implementing the following computational paradigm:
\begin{equation}
    \left\{ \begin{array}{l}
        Q=XW_{\text{q}},\ K=XW_{\text{k}},\ V=X\\
        S_W = \text{softmax}\left(\frac{Q K^T}{\sqrt{d_k}}\right)V\\
    \end{array} \right. ,
    \label{con:SCAM1}
\end{equation}
where $W_q$ and $W_k$ represent the learnable weights, and $d_k$ denotes the number of channels in the input feature $X$.

Beside constructing global channel dependence using standard convolution $\text{Conv}_{1\times 1}$, we introduce 1D convolution $f_{1\text{D}}$ to achieve localized inter-channel attention, given as follows:
\begin{equation}
    F_{\text{loc}}=f_{1\text{D}}\left( f_{\text{GAP}}\left( \text{Conv}_{1\times 1}\left( X \right) \right) \right) .
    \label{con:SCAM2}
\end{equation}
Here, $f_\text{GAP}$ represents the Global Average Pooling.

Subsequently, global channel attention $C_W$ is established by applying a 1×1 convolution to the local feature $F_{\text{loc}}$. Finally, by combining the joint spatial feature $S_W$ from the spatial attention, the output feature $F_\text{SCA}$ is computed through an element-wise multiplication operation $\otimes$ as follows:
\begin{equation}
    F_{\text{SCA}}=\text{Conv}_{1\times 1}\left( F_{\text{loc}} \right) \otimes S_W
    \label{con:SCAM3}
\end{equation}

\begin{figure}[t]
    \centering
    \includegraphics[width=0.75\linewidth]{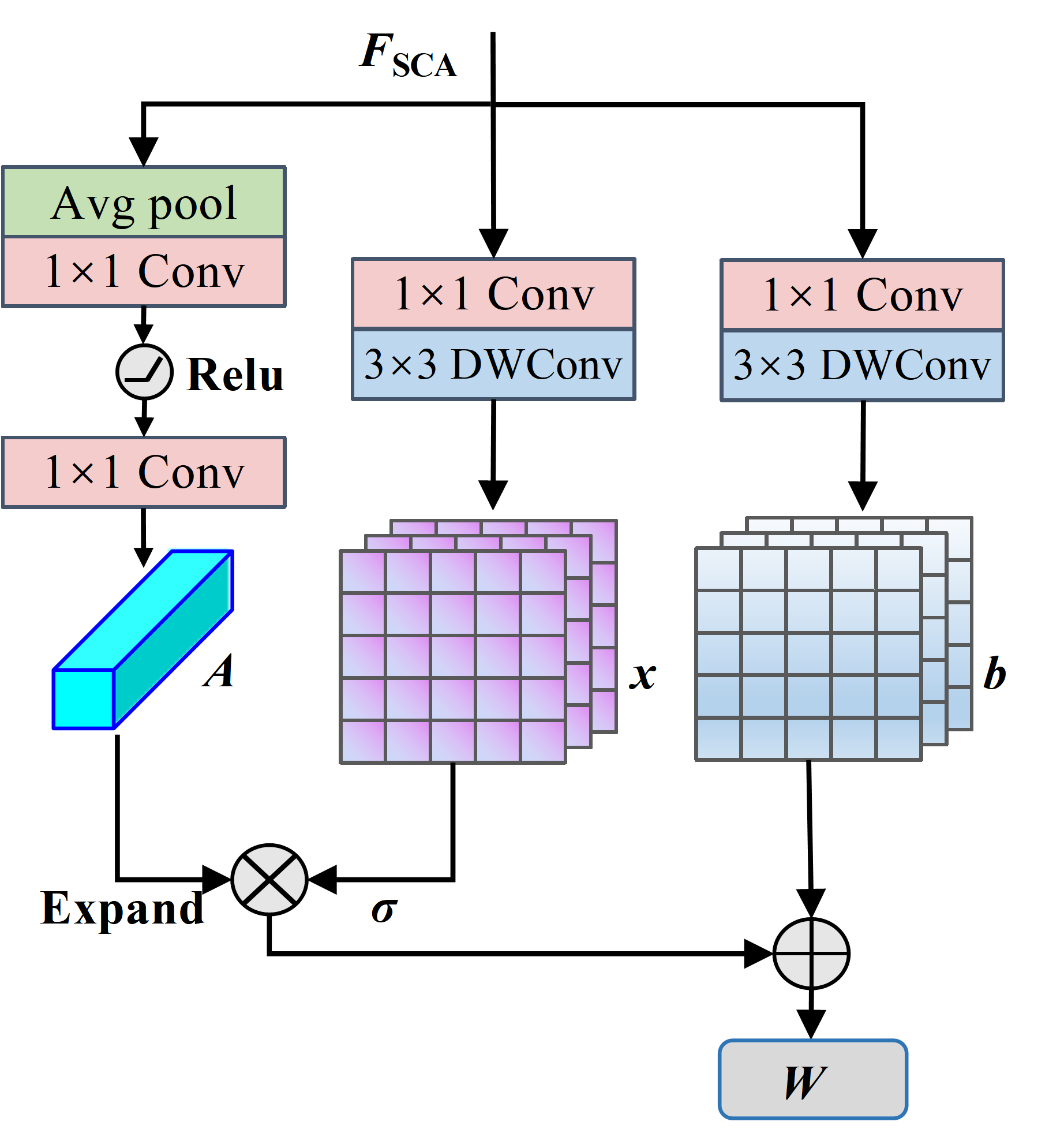}
    \caption{The proposed weight decoupling module. Generating affine transform factors $A$, $x$, and $b$ by three branches and generating fusion weight $W$ using multiplication operation $\otimes$ and summation operation $\oplus$.}
    \label{fig:WDM}
\end{figure}

\textbf{Weight Decoupling Module:} Inspired by the observation that identical features should exhibit consistency in both spatial and channel weight distributions, we introduce a novel weight generation mechanism to establish a fine-grained relationship between the goal and observed features. As illustrated in Fig.~\ref{fig:WDM}, the proposed Weight Decoupling Module (WDM) constructs spatial and channel attention through three distinct branches, utilizing affine transformation theory to dynamically generate adaptive weights.

For the channel dimension, we employ a mechanism inspired by the conventional channel attention approach to generate the weight feature $A$. Specifically, the global feature is first extracted using global average pooling $f_\text{GAP}$, followed by two consecutive 1×1 convolution layers to generate channel weight:	
\begin{equation}
    A = \text{Conv}_{1 \times 1}  \left( \delta \left( \text{Conv}_{1 \times 1} \left( f_\text{GAP}(F_\text{SCA}) \right) \right ) \right).
    \label{con:WDM1}
\end{equation}
Here, $\delta$ is the Relu activation function, and $F_\text{SCA}$ represents the output feature of SCA.

Simultaneously, feature transformation is achieved through 1×1 convolution, while spatial weight $x$ is captured by 3×3 depthwise convolution $\text{DWConv}_{3 \times 3}$ and a sigmoid function $\sigma$:
\begin{equation}
    x = \sigma \left( \text{DWConv}_{3 \times 3}  \left( \text{Conv}_{1 \times 1}(F_\text{SCA}) \right) \right).
    \label{con:WDM2}
\end{equation}
Finally, weights from different branches are fused as follows:
\begin{equation}
    W=A\cdot x+b.
    \label{con:WDM3}
\end{equation}

\subsubsection{Self-distillation}
\begin{figure}[t]
    \centering
    \includegraphics[width=0.9\linewidth]{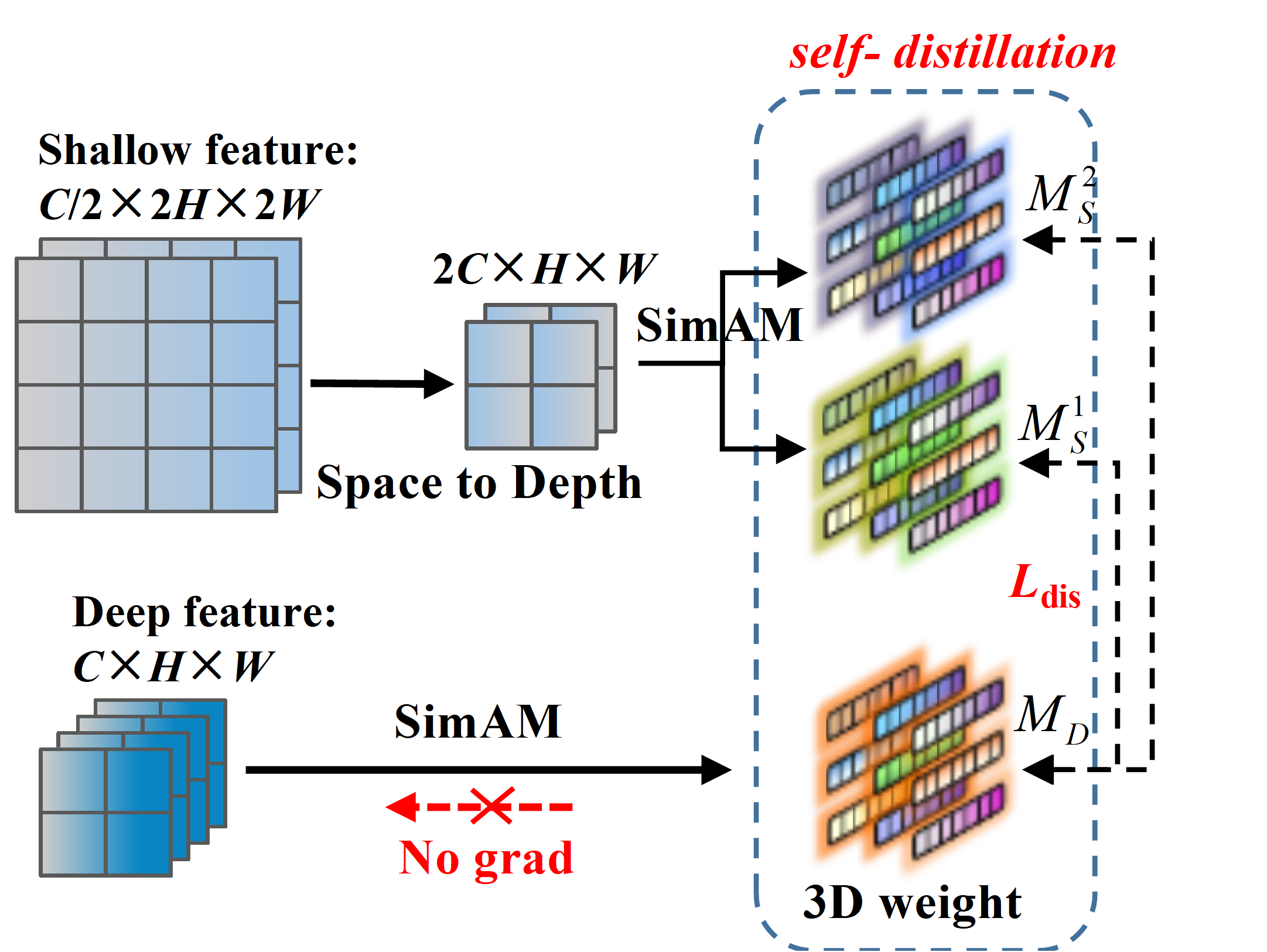}
    \caption{The self-distillation mechanism. The spatial features of the shallow feature are first transformed to the channel by space-to-depth operation, then the parameter-free attention mechanism SimAM \cite{yang2021simam} is introduced to generate the 3D weights, and finally the model is supervised by the similarity between the 3D weights.}
    \label{fig:loss}
\end{figure}

Although ResNet9 achieves efficient feature extraction by stacking multiple residual blocks, its performance is inherently limited by the model's shallow depth. To overcome this constraint, we introduce a guidance mechanism that allows deep features to supervise shallow features, enabling the shallow network to approximate the weight distributions of the deeper network \cite{zhang2019your,9009567}. As illustrated in Fig.~\ref{fig:loss}, the proposed self-distillation mechanism adheres to two core principles: (1) The distillation pipeline is entirely parameter-free, leveraging only the intrinsic properties of the features themselves. (2) Deep features are treated as ground truth for supervising shallow features, and gradients from the deep network are truncated during backpropagation to prevent interference. This approach enhances the shallow network's ability to capture richer representations while maintaining computational efficiency.

To match the dimensions between the shallow feature $I \in \mathbb{R}^{C/2\times 2H\times 2W}$ and deep feature $I_d \in \mathbb{R}^{C\times H\times W}$, we employ a space-to-depth transformation \cite{khanam2024yolov5} to transform  $I$ to $I_s\in \mathbb{R}^{2C\times H\times W}$ as follows: 
\begin{equation}
    I_s\left[ i,j,c \right] =I\left[ 2i+\delta _i,2j+\delta _j, \frac{c}{4}  \right] ,\ i,j\in \{1,\cdots ,\frac{H}{2}\},
    \label{con:loss1}
\end{equation} 
where $\left( \delta _i,\delta _j \right) \in \left\{ \left( 0,0 \right) ,\left( 0,1 \right) ,\left( 1,0 \right) ,\left( 1,1 \right) \right\} $.

Subsequently, the shallow feature $I_s$ is decomposed into two feature $ I_{s}^{1},I_{s}^{2}\in \mathbb{R}^{C\times H\times W} $ along the channel dimension. Corresponding weight $M$ is then generated through the parameter-free SimAM attention mechanism \cite{yang2021simam}, given as follows:	
\begin{equation}
    N_c = 4 \left( \frac{\sum_{i,j} D_{c,i,j}}{H \cdot W - 1} + \lambda \right),
    M_{c,i,j} = \sigma \left( \frac{D_{c,i,j}}{N_c} + 0.5 \right),
    \label{con:loss2}
\end{equation}
where $D$ is the squared difference of the feature $I_d, I_{s}^{1},I_{s}^{2}$, and $\lambda$ is a regularization parameter, set to a default value of 1e-4. Here $N_c$ and $M_c$ represent the normalization term for the denominator and attention weight of the $c$-th channel, respectively.

Finally, to quantify the alignment between the deep attention feature $M_D$ and shallow attention feature $M_S$
, we utilize cosine similarity. This similarity measure is used to construct the distillation loss $L_{dis}$, given as follows:
\begin{equation}
    L_{dis} = 0.2 \times (L_{C1} + L_{C2}), L_C = 1-\frac{M_S \cdot M_D}{\|M_S\| \|M_D\|}.
    \label{con:loss3}
\end{equation}

\subsection{Image Scene Graph Module}\label{sec2}

\begin{figure*}[t]
    \centering
    \includegraphics[width=0.9\linewidth]{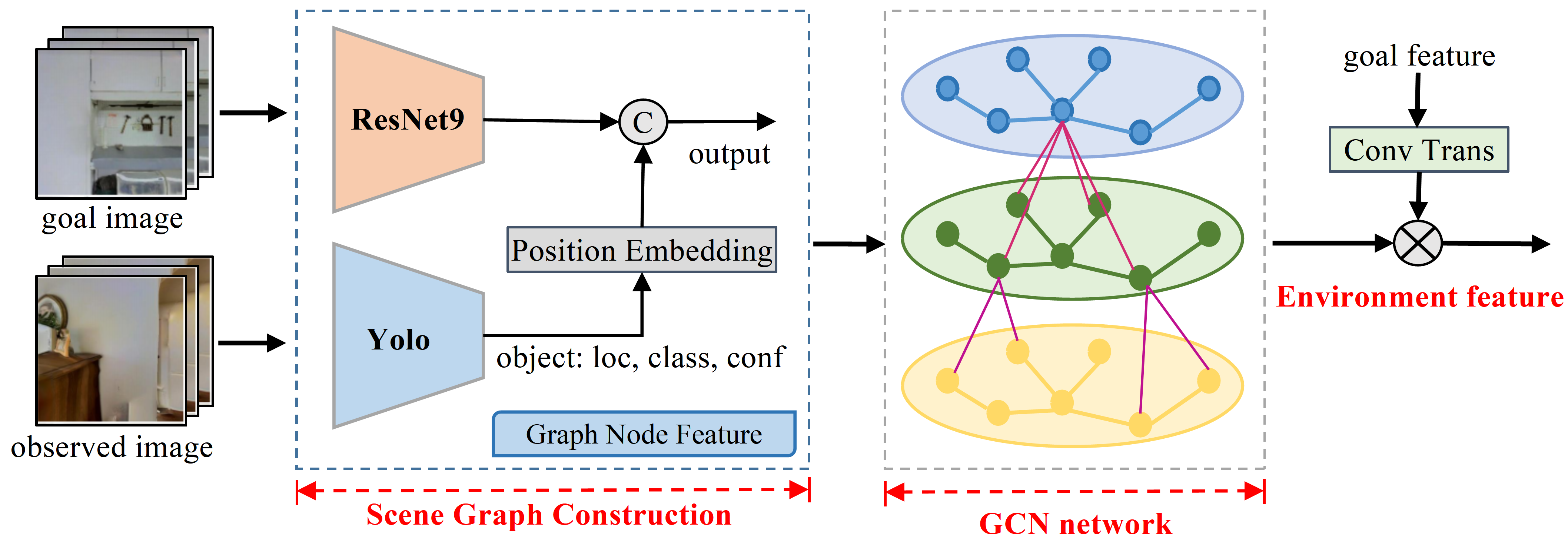}
    \caption{Scene graph architecture for capturing environmental information. First, the image feature and instance objects are extracted by pre-trained ResNet9 and YOLOv5, respectively, and the location, class, and confidence of the instance objects are embedded into the image feature with position encoding, which constitutes the node feature of the scene graph. Subsequently, an adjacency matrix is constructed from the cosine similarity of the graph node features, and a Graph Convolutional Neural (GCN) network is constructed to extract the scene graph feature. Finally, the goal feature is introduced to select the important scene feature.}
    \label{fig:graph}
\end{figure*}

Beyond the cognition of the goal and current observations, navigation requires a certain understanding of the explored environmental information to enable a more efficient goal search. Inspired by the 3D scene graph~\cite{armeni20193d}, we propose the Image Scene Graph Module, as illustrated in Fig.~\ref{fig:graph}. Since our input is limited to images, we focus on the association between different images and instance objects and embed the optimization of the scene graph into the model training. We use image and instance features as node features of the scene graph, through the similarity between node features as the value of the adjacency matrix, and then introduce a graph convolutional neural network to aggregate the environment information.

Specifically, we employ another pre-trained ResNet9 to extract image feature $F_{\text{Image}}$ and utilize a YOLOv5~\cite{khanam2024yolov5} to extract instance features $F_{\text{Instance}}$, which include the position $O_{\text{loc}}$, class $O_{\text{class}}$, and confidence $O_{\text{conf}}$. Specifically, $O_{\text{loc}}=\left( \frac{x_1}{W},\frac{y_1}{H},\frac{x_2}{W},\frac{y_2}{H} \right) $, where $(W,H)$ denote the width and height of the image, and $(x_1,y_1,x_2, y_2)$ represents the bounding box coordinates of the target in the image. For the class information, we define $\ O_{\text{class}}=\log \left( c \right)$, where $c$ is the target class index. Since $c$ is typically a large natural number, we constrain it to a logarithmic space to ensure numerical stability. We concatenate $f_{\text{Cat}}$ these features and  use a Multilayer Perceptron (MLP) to map them into the instance feature as follows:
 \begin{equation}
F_{\text{Instance}}= \ \text{MLP}\left( f_\text{Cat}(O_{\text{loc}},O_{\text{class}},O_{\text{conf}}) \right) .
    \label{con:graph1}
\end{equation}
The scene graph node feature is concatenated by the image feature and instance feature, given as 
\begin{equation}
F_{\text{Node}}=f_\text{Cat}\left( F_{\text{Image}}, F_{\text{Instance}} \right) .
    \label{con:graph1}
\end{equation}

Based on the scene graph node features $F_{\text{Node}}$, we construct the adjacency matrix $G$ by calculating the cosine similarity between the node features, thus constructing the image scene graph, given as follows: 
\begin{equation}
    G_{i,j} = \frac{F_{\text{Node}}{_i} \cdot F_{\text{Node}}{_j}}{\|F_{\text{Node}}{_i}\| \|F_{\text{Node}}{_j}\|}, G \in \mathbb{R}^{N\times N},
    \label{con:graph3}
\end{equation}
where the values of the adjacency matrix $G_{i,j}$ represent the similarity between each pair of node features. 

Finally, we encode the scene graph features using a three-layer graph convolutional neural network and gradually compress the node feature dimension through average pooling, given as follows:	
\begin{equation}
    H^{(l+1)} = \delta \left(\hat{G} H^{(l)} W^{(l)}\right),H_{\text{graph}} = \text{mean}(H^{(L)}),
    \label{con:graph4}
\end{equation}
where $\hat{G}$ and $H$ are the adjacency and feature matrices of the scene graph, $W$ is the weight of the graph convolutional neural network, and $\delta$ is the Relu activation function.

Concurrently, we fuse the goal feature $F_{\text{goal}}$ with the scene graph information $H_{\text{graph}}$ to capture environmental information $F_{\text{env}}$ that is critical for the goal, given as follows:
\begin{equation}
    F_{\text{env}} = \text{Conv}(F_{\text{goal}}) \otimes H_{\text{graph}}.
    \label{con:graph5}
\end{equation}

\subsection{Navigation Policy}\label{sec3}
Based on the output feature of the backbone \( F_{\text{backbone}} \) and environmental feature \( F_{\text{env}} \), we train the navigation policy \( \pi \) using reinforcement learning, given as follows:	
\begin{equation}
    a_t \sim \pi(f_\text{Cat}(F_{\text{backbone}},F_{\text{env}}) \oplus a_{t-1} \mid h_{t-1}).
\end{equation}
Here, \( \pi \) is the navigation policy for the Actor-Critic architecture, and \( h_{t-1} \) represents the previous hidden state of the Gate Recurrent Unit (GRU). Following the previous methods \cite{sun2023fgprompt,al2022zero}, we combine the previous action \( a_{t-1} \) to predict the next state value \( c_t \) and action \( a_t \), and conduct end-to-end training using the Proximal Policy Optimization (PPO) algorithm \cite{schulman2017proximal}.

\section{Experimental Evaluation}
\label{sec:exp}

We present our experiments to demonstrate the performance of our method. Here, we make four core claims, which are:
(i) Our navigation algorithm achieves state-of-the-art (SOTA) navigation performance and exhibits strong generalization capabilities.
(ii) The proposed feature fusion architecture generates fine-grained weights for fusion, significantly enhancing feature representation.
(iii) Deep features guide the shallow network to learn the weights of the deep network, thereby improving the feature extraction performance of the backbone network.
(iv) Our proposed image scene graph effectively encodes surrounding environmental information, leading to improved navigation performance.

These claims are supported by extensive experimental results, which validate the effectiveness and robustness of our approach in various navigation scenarios.

\subsection{Experimental Setup}

\textbf{Dataset and Metrics: }In image-goal navigation, we implement the pipeline using the Habitat simulator~\cite{savva2019habitat,szot2021habitat} and utilize the Gibson dataset~\cite{mezghan2022memory}, which comprises 72 training scenes and 14 testing scenes. Furthermore, we employ the test episodes from HM3D~\cite{al2022zero} to further validate the performance of our method in new scenes. For evaluation metrics, we use SR (Success Rate) and SPL (Success weighted by Path Length) to comprehensively assess the navigation performance. SR measures the proportion of the robot successfully reaches the goal location in navigation tasks. SPL is a more rigorous metric, as it not only considers whether the robot successfully reaches the target but also takes into account the path length traveled by the robot during navigation. The number of steps executed in Section~\ref{EXP1} and Section~\ref{EXP2} is 100\,M, which is designed to quickly evaluate the different components. In the comparative experiment in Section \ref{EXP3}, we follow the baseline~\cite{sun2023fgprompt} setup and set the step number as 500\,M.

\subsection{Comparison with Mainstream Methods}\label{EXP3}

\textbf{Experiment on Gibson: }As shown in Tab.~\ref{tab:EXP3_1}, the performance of various methods on the Gibson dataset shows several key points. Specifically, methods based on ResNet50, such as ZSON \cite{majumdar2022zson} and OVRL \cite{yadav2023offline}, outperform ZER \cite{al2022zero}, which is built upon ResNet9. Moreover, certain approaches like NTS \cite{chaplot2020neural}, which rely on both visual and inertial information for image-goal navigation, although achieving relatively superior performance, are highly dependent on sensor data. Notably, our proposed method, which solely depends on a lightweight ResNet9 as the backbone and takes a single RGB image as input, surpasses mainstream methods in both SPL and SR metrics. To be precise, our method achieves an SPL of 67.8\%, representing a substantial improvement of 5.4\% over the baseline model FGPrompt (62.4\%), and attains an SR of 91.0\%. This shows that through refined feature-guided fusion and scene graph enhancement, our approach achieves higher navigation success and efficiency, which can efficiently accomplish goal-driven visual navigation tasks. 

\begin{table}[t]
    \centering
    \caption{Model training and validation on Gibson dataset. \dag\enspace is the reproduced results using the open-source code and weights.}
    \label{tab:EXP3_1}
    \setlength{\tabcolsep}{5pt}
    \begin{tabular}{lcccc}
        \hline
        Method & Backbone & Sensor & SPL & SR \\ \hline
        NTS \cite{chaplot2020neural} & ResNet9 & RGBD+Pose & 43.0\% & 63.0\% \\ 
        Acti-Neur-SLAM \cite{chaplot2020learning} & ResNet9 & RGB+Pose & 23.0\% & 35.0\% \\ 
        SPTM \cite{savinov2018semi} & ResNet9 & RGB+Pose & 27.0\% & 51.0\% \\
        ZER \cite{al2022zero} & ResNet9 & RGB & 21.6\% & 29.2\% \\ 
        ZSON \cite{majumdar2022zson} & ResNet50 & RGB & 28.0\% & 36.9\% \\ 
        OVRL \cite{yadav2023offline} & ResNet50 & RGB & 27.0\% & 54.2\% \\ 
        OVRL-V2 \cite{yadav2023ovrl} & ViT-Base & RGB & 58.7\% & 82.0\% \\ 
        FGPrompt\dag \enspace\cite{sun2023fgprompt} & ResNet9 & RGB & 62.4\% & 90.4\% \\ 
        \textbf{Ours} & \textbf{ResNet9}  & \textbf{RGB} & \textbf{67.8\%}  & \textbf{91.0\%}  \\ \hline
    \end{tabular}
\end{table}

\textbf{Generalziation on HM3D: }To further validate cross-scene performance, the weight trained on Gibson is directly evaluated on HM3D without any fine-tuning, as shown in Tab.~\ref{tab:EXP3_2}. Compared to the baseline model FGPrompt, our method achieves a commendable navigation success rate and attains the highest SPL of 42.7\%. This shows that our approach can balance generalizability and navigation performance in new scenes, and reduce unnecessary detours to improve overall navigation efficiency.

\begin{table}[t]
    \setlength{\tabcolsep}{13pt}
    \centering
    \caption{The model is trained on the Gibson dataset and validated on the HM3D dataset.}
    \label{tab:EXP3_2}
    \begin{tabular}{lccc}
        \hline
        Method & Backbone & SPL & SR \\ \hline
        Mem-Aug \cite{mezghan2022memory} & ResNet18 & 3.5\% & 1.9\% \\ 
        NRNS \cite{hahn2021no} & ResNet18 & 4.2\% & 6.6\% \\
        ZER \cite{al2022zero} & ResNet9 & 6.3\% & 9.6\% \\ 
        FGPrompt\dag \enspace\cite{sun2023fgprompt} & ResNet9 & 38.6\% & 72.7\% \\ 
        \textbf{Ours} & \textbf{ResNet9} & \textbf{42.7\%} & \textbf{73.4\%} \\ \hline
    \end{tabular}
\end{table}

\begin{figure}[t]
    \centering
    \includegraphics[width=1\linewidth]{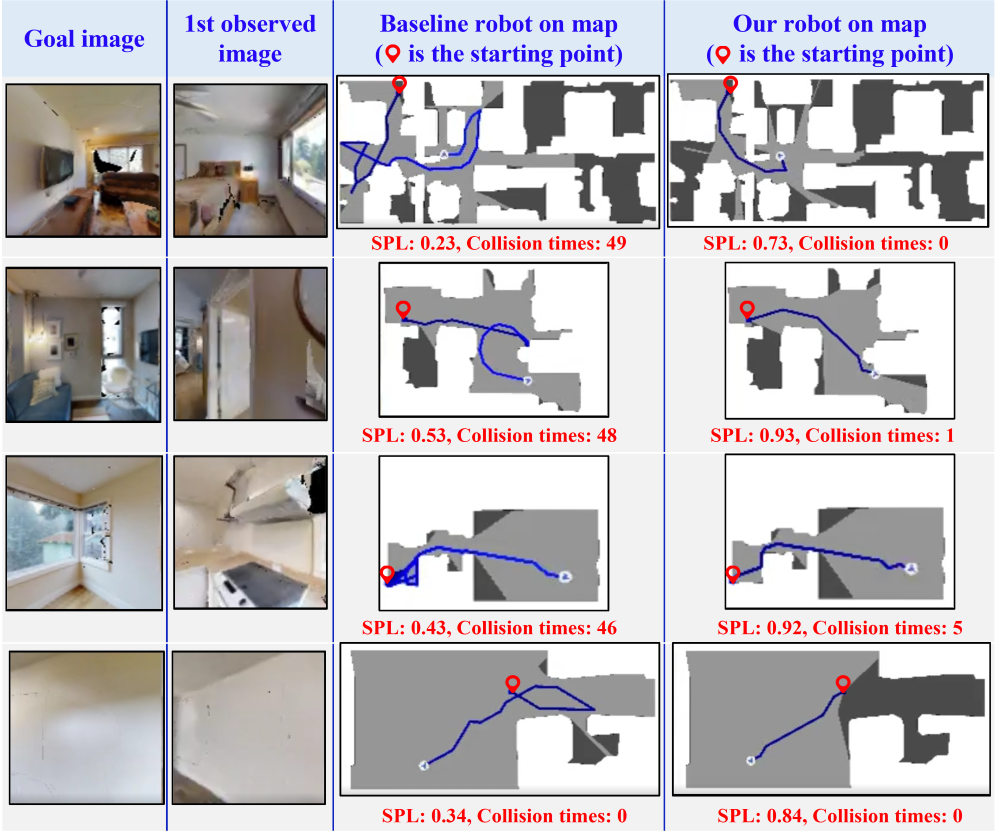}
    \caption{The navigation path of the robot on the ground-truth map. The “blue line” is the robot's navigation path.}
    \label{fig:map}
\end{figure}

\textbf{Navigation on Ground-truth Map: }As illustrated in Fig.~\ref{fig:map}, the navigation paths generated by the robot in different scenarios demonstrate a clean and visualized point. Our approach can balance navigation success and efficiency, which further demonstrates the effectiveness of the proposed refined feature guidance and scene graph strategy. Compared to the baseline method, our approach reduces the number of unnecessary bypasses and collisions thus improving the navigation efficiency.

\subsection{Ablation Experiment of Our Method}\label{EXP2}

The contributions of the proposed different components to model performance are shown in Tab.~\ref{tab:EXP2}. When the proposed feature fusion module replaces FilM of Baseline, the model's SPL and SR reach 59.30\% and 83.52\%, respectively, both increasing by at least about 7\%. This indicates that the spatial-channel attention effectively models local and global information and combines with the affine transformation of the weight decoupling module to finely process feature associations, which can more effectively capture cross-modal feature associations. Meanwhile, the proposed self-distillation mechanism guides shallow networks to learn deep networks, significantly improving model performance without adding additional learnable parameters. Modeling environmental information at different moments through the scene graph significantly improves planning performance. In summary, the proposed components enhance feature utilization, significantly improving the success rate of image-goal navigation.

\begin{table}[t]
    \setlength{\tabcolsep}{8pt}
    \centering
    \caption{Ablation studies on our designed components.}
    \label{tab:EXP2}
    \begin{tabular}{ccc|cc} 
        \hline
         \multicolumn{3}{c|}{Components} & SPL & SR \\ \cline{1-3}
        Fusion & Distillation & Scene Graph &  &  \\ \hline
        - & - & - & 52.35\% & 75.21\% \\
        \checkmark & - & - & 59.30\% & 83.52\% \\
        \checkmark	& \checkmark & - & 60.45\% & 85.69\% \\
        \checkmark & \checkmark & \checkmark & \textbf{64.20\%} & \textbf{87.45\%} \\ \hline
    \end{tabular}
\end{table}

\subsection{Analysis of Fusion Strategy}\label{EXP1}

\textbf{Fusion Architecture Validation: }In order to verify the effectiveness of the proposed feature fusion architecture, FilM~\cite{perez2018film} is used as the baseline for ablation experiments. As shown in Tab.~\ref{tab:EXP1_1}, the results indicate that the complete fusion architecture outperforms models with any component removed in terms of both SPL and SR. When feature fusion is performed without learnable parameters ${W_1,W_2}$, the model's SPL and SR significantly decrease by approximately 4.5\%. After removing the $W_1$, the model is unable to learn weights from the goal feature, leading to a significant decline in navigation success rate. Similarly, when the observed features are not adjusted by weight $W_2$, the model's performance also degrades. In summary, each component of the feature fusion architecture plays an important role in navigation performance.

\begin{table}[!h]
    \setlength{\tabcolsep}{20pt}
    \centering
    \caption{Ablation experiments with feature fusion architecture.}
    \label{tab:EXP1_1}
    \begin{tabular}{lcc}
        \hline
        Fusion Architecture & SPL & SR \\ \hline
        \textbf{FilM~\cite{perez2018film}} & \textbf{52.35\%} & \textbf{75.21\%} \\
        FilM w/o $W_1$ & 51.73\% & 69.76\% \\
        FilM w/o $W_2$ & 49.29\% & 69.17\% \\
        FilM w/o $W_1,W_2$ & 47.95\% & 70.57\% \\ \hline
    \end{tabular}
\end{table}

\textbf{Spatial-Channel Attention: }Based on the Baseline architecture, we performed ablation experiments on the proposed spatial-channel attention (SCA) to verify the necessity of focusing on features at different levels, as shown in Tab.~\ref{tab:EXP1_2}. The results show that by introducing spatial and channel attention \cite{woo2018cbam} to focus on features (CAM and SAM), the navigation success rate significantly increased. It is worth noting that by using the proposed SCA for focusing salient features, the model's SPL and SR reached 57.77\% and 83.21\%, respectively. Considering the navigation success rate and efficiency, the proposed SCA is superior to the method using only spatial or channel attention. In addition, in terms of the connection method of spatial and channel features, compared to simple serial or parallel methods, the proposed module fully utilized channel weights to focus on the important channels of spatial features, effectively associating local and global information.
\begin{table}[t]
    \setlength{\tabcolsep}{19pt}
    \centering
    \caption{Ablation and comparison experiments of SCA.}
    \label{tab:EXP1_2}
    \begin{tabular}{lcc}
        \hline
        Feature Focus Strategy & SPL & SR \\ \hline
        Baseline & 52.35\% & 75.21\% \\
        CAM \cite{woo2018cbam} & 55.57\% & 80.26\% \\
        SAM \cite{woo2018cbam} & 51.65\% & 69.57\% \\
        SCA w/ serial & 57.48\% & 83.12\% \\
        SCA w/ Parallel\&add & 57.79\% & 80.38\% \\
        SCA w/o channel & \textbf{57.97\%} & 79.90\% \\
        SCA w/o spatial & 56.48\% & 80.14\% \\
        \textbf{SCA (Ours)} & 57.77\% & \textbf{83.21\%} \\
        \hline
    \end{tabular}
\end{table}

\textbf{Weight Decoupling Module: } The WDM generates and fuses weights in the channel and spatial dimensions, aiming to exploit finer features to improve model performance. The experimental results are shown in Tab.~\ref{tab:EXP1_3}. Compared with the convolution operation of Baseline, after introducing the WDM, the SR remained almost unchanged (up to 74.95\%), while the SPL increased by 2.84\% (up to 55.19\%). Simultaneously, in the ablation experiments of the affine transformation weight generation strategy $y=Ax+b$, each component $(A, x, b)$ played an important role in the feature fusion. In summary, with a refined weight generation strategy, WDM effectively captures the association between goal and observation features.

\begin{table}[t]
    \setlength{\tabcolsep}{16pt}
    \centering
    \caption{Ablation experiment for WDM}
    \label{tab:EXP1_3}
    \begin{tabular}{lcc}
        \hline
        Weight Generation Strategy & SPL & SR \\ \hline

         Baseline & 52.35\% & \textbf{75.21}\% \\
         WDM w/o $A$ & 51.19\% & 71.14\% \\
         WDM w/o $x$ & 52.36\% & 73.10\% \\
         WDM w/o $b$ & 47.98\% & 75.14\% \\ 
         \textbf{WDM (Ours)} & \textbf{55.19\%} & 74.95\% \\
         \hline
    \end{tabular}
\end{table}

\subsection{Runtime Analysis}

The runtime impact of different components in our method is summarized in Tab.~\ref{tab:runtime}. On average, our method achieves a runtime of 18.7\,ms on a GPU and 35.2\,ms on a CPU, namely 53.5\,FPS (Frames Per Second) and 28.4\,FPS, respectively. These results satisfy real-time operational requirements, making our proposed approach well-suited for real-world mobile robotics applications.

\begin{table}[h]
    \setlength{\tabcolsep}{3pt}
    \centering
    \caption{Runtime analysis of our method.}
    \begin{tabular}{ccccc}
    \hline
    Fusion & Distillation & Scene Graph & RTX 3080 (GPU) & i9-12900 (CPU) \\ \hline
    - & - & - & 1.8 ms & 4.3 ms \\ 
    \checkmark & - & - & 3.6 ms & 7.3 ms \\ 
    \checkmark & \checkmark & - & 3.6 ms  & 7.3 ms \\ 
    \checkmark & \checkmark & \checkmark & 18.7 ms & 35.2 ms \\ \hline
    \end{tabular}
    \label{tab:runtime}
\end{table}

\section{Conclusion}
\label{sec:conclusion}
In this paper, we propose a novel image-goal navigation approach, RFSG. Our approach mines fine-grained associations between the goal and the observation within limited image information while maintaining a lightweight architecture. We introduce a novel feature fusion module that enables the network to focus on salient goal features from different dimensions. By incorporating a self-distillation mechanism, we guide the shallow network to mimic the weights of the deep network in a parameter-free manner, further enhancing feature representation. Additionally, we propose to introduce an instance scene graph to better encode scene environmental information, significantly improving navigation performance and efficiency.

Experiments on different datasets demonstrate that our method achieves state-of-the-art success rates and path efficiency among baseline methods, with an inference speed of up to 53.5\,FPS on an RTX 3080 GPU. Moreover, our approach exhibits excellent generalization capabilities, enabling seamless adaptation to new scenes without any adjustments. To validate the effectiveness of each component, we conduct comprehensive ablation studies, which confirm the contributions of all proposed modules to the overall performance. These results highlight the robustness, efficiency, and scalability of our proposed method in navigation scenarios.


\bibliographystyle{ieeetr}

\bibliography{glorified,new}

\end{document}